\documentclass{article}
\usepackage{genai4health_2026}
\usepackage[utf8]{inputenc}
\usepackage[T1]{fontenc}
\usepackage{microtype}
\usepackage{amsmath,amssymb,amsthm}
\usepackage{booktabs,tabularx}
\usepackage{graphicx}
\usepackage{wrapfig}
\usepackage{placeins}
\usepackage{tcolorbox}
\usepackage{float}
\usepackage{needspace}
\newtcolorbox{promptbox}[1]{
  colback=black!2,
  colframe=black!65,
  colbacktitle=black!65,
  coltitle=white,
  title={#1},
  fonttitle=\small\bfseries,
  fontupper=\small,
  boxrule=0.45pt,
  arc=2mm,
  boxsep=1mm,
  left=2mm,
  right=2mm,
  top=1.5mm,
  bottom=1.5mm,
  toptitle=0.4mm,
  bottomtitle=0.4mm,
  before skip=0pt,
  after skip=0pt
}
\newcommand{\promptfield}[1]{\par\smallskip{\sffamily\bfseries\color{black!65}[#1]}\par\nobreak\smallskip}
\newcommand{\promptdivider}{\par\medskip\hrule height 0.35pt\relax\medskip}
\newcommand{\promptslot}[1]{\texttt{\{#1\}}}

\usepackage[ruled,vlined,linesnumbered]{algorithm2e}
\usepackage{url}
\usepackage[colorlinks=true,linkcolor=blue,citecolor=blue,urlcolor=blue]{hyperref}
\hypersetup{pdftitle={CausalEvo: Causal Skill Self-Evolution for Health LLM Agents},pdfauthor={},pdfsubject={GenAI for Health Workshop at NeurIPS 2026, Sydney},pdfkeywords={health LLM agents, skill evolution, skill intervention, reliability evaluation}}
\theoremstyle{definition}
\newtheorem{proposition}{Proposition}

\newcommand{\E}{\mathbb{E}}

\newcommand{\one}{\mathbf{1}}
\newcommand{\method}{CausalEvo}
\title{CausalEvo: Causal Skill Self-Evolution\\for Health LLM Agents}
\author{%
  Yuchen Zhang \\
  The University of Texas Health Science Center at Houston \\
  \texttt{yuchen.zhang@uth.tmc.edu}
}
\hypersetup{pdfauthor={Yuchen Zhang}}

\date{}

\begin{document}
\maketitle
\begin{abstract}
Large language model (LLM) agents for health can accumulate reusable textual skills from task feedback, such as calculation criteria, measurement time points, and formula units. Deciding which skills to retain is consequential: useful rules improve later answers, while faulty generalizations propagate errors. We propose \method{}, a method for causal skill self-evolution that estimates the relative effects of deliberately changed skill sets while holding the model weights, feedback cases, and rule count fixed. At each evolution step, a frozen LLM proposes candidate rules, which are pooled with the incumbent skills so that the current skill set remains a feasible choice. Paired skill replacements against shared background rules are scored on the same feedback cases; least squares converts the score differences into relative skill values, and the top-ranked rules are retained, favoring incumbents on ties. A decision-loss bound separates candidate support, approximation, and estimation errors. On six MedCalc-Bench-Verified tasks with 120 held-out cases, \method{} achieves 60.00\% accuracy, compared with 55.83\% for SkillOpt and 51.67\% without skills. Across the recorded training configurations, which use different optimization budgets, it consumes 55.6\% fewer training tokens than SkillOpt. Gains concentrate in particular calculations alongside regressions on others, and paired accuracy intervals include zero. \method{} turns task feedback into controlled decisions about skill retention and replacement. Project repository: \url{https://github.com/txbrook/CausalEvo}.
\end{abstract}

\section{Introduction}
\label{sec:intro}
Health LLM agents with frozen weights can retain instructions distilled from feedback. In medical calculation, these skills may specify a scoring criterion, a measurement time point, or a formula's units. A useful rule can improve later answers; an incorrect generalization can also persist across cases. This makes the decision to retain or replace a skill part of agent reliability: what evidence should guide skill evolution?

External-memory adaptation stores knowledge acquired from feedback \citep{reflexion,voyager,ace}, and textual optimizers already combine proposals with evaluation \citep{gepa,skillboost,cass}. We focus on how changing the supplied skill set affects task performance.

\method{} treats skill evolution as a causal comparison: on the same feedback case, we change the supplied rules while keeping the model weights and rule count fixed, and observed score differences estimate the relative effects of the compared skill sets. We call this controlled change a skill intervention. Our organizing principle is that \textbf{skill selection should use comparisons that preserve the deployed rule capacity}. A generator that rewrites every candidate may remove an old skill before retention can be evaluated, so the incumbent skill stays in the feasible set. Comparisons that change the number of rules mix content effects with size effects, so compared skills have equal capacity.

\begin{figure}[t]
\centering
\includegraphics[width=\linewidth]{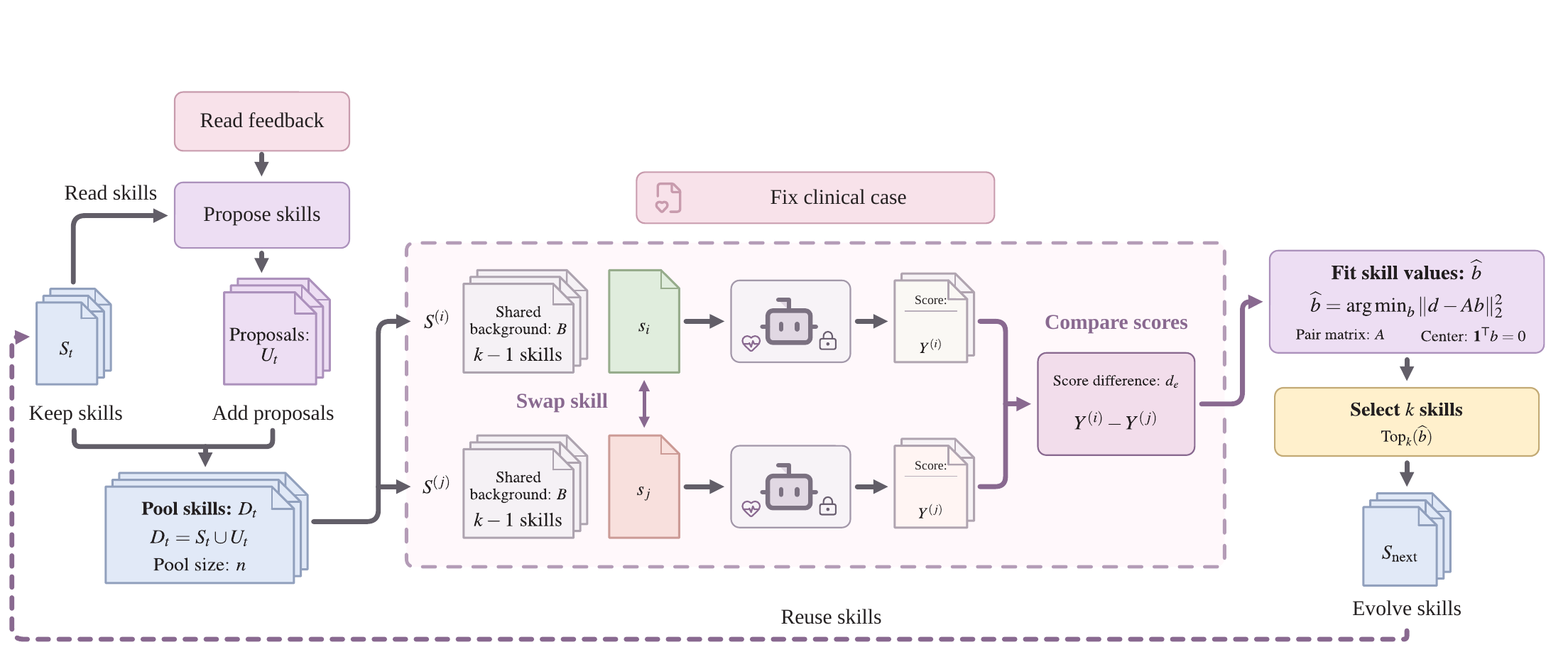}
\caption{Overview of \method{}. Incumbent skills ($S_t=M_t$) and model-generated proposals form a candidate pool. On a fixed feedback case, two skill sets share $k-1$ background rules and exchange one rule. The frozen actor scores both skill sets; paired score differences fit relative skill values for top-$k$ selection, with ties favoring incumbent rules.}
\label{fig:interface}
\end{figure}

The frozen model proposes rules; a program collects paired exchanges on a comparison graph, estimates relative rule values, and selects the next skill (Figure~\ref{fig:interface}). Proposal quality remains the model's responsibility; credit estimation and selection are programmatic.

Our contributions are (i) a formulation of skill evolution with incumbent-containing feasible sets; (ii) a fixed-capacity exchange estimator and decision-loss decomposition, using classical statistical tools to analyze textual skills; and (iii) a 120-case held-out comparison with SkillOpt and no skill, learning dynamics, output-validity diagnostics, and a credit-assignment ablation.

\section{Related work}
\label{sec:related}

\paragraph{Continual learning.}
Elastic Weight Consolidation protects parameters important to previous tasks \citep{ewc}; Gradient Episodic Memory constrains updates using stored examples and distinguishes accuracy from backward and forward transfer \citep{gem}. External-memory methods retain verbal reflections (Reflexion), executable skills (Voyager), or incrementally curated playbooks (ACE) \citep{reflexion,voyager,ace}. \method{} studies retention with frozen parameters and a finite textual skill. The incumbent remains a feasible decision, and we assess retention through task-specific backward transfer.

\paragraph{Textual optimization.}
Self-Refine revises individual outputs through self-feedback \citep{selfrefine}; TextGrad propagates textual feedback through computation graphs \citep{textgrad}; GEPA tests reflected prompt changes and combines complementary candidates \citep{gepa}. SkillOpt optimizes external skill documents through feedback and validation \citep{skillopt}. SkillBoost evaluates repairs under regression constraints \citep{skillboost}, while SkillCAT combines contrastive extraction, patch assessment, and skill topology \citep{skillcat}. \method{} shares proposal generation and empirical evaluation, focusing on comparisons among deployable skills in an incumbent-containing, fixed-capacity family. We compare with adapted SkillOpt; broader end-to-end comparisons remain open.

\paragraph{Composition-aware valuation.}
SkillShapley attributes performance to skill steps using Shapley values and boundary-adaptive sampling \citep{skillshapley}. CASS combines paired coalition-knockout margins with aggregate reward when accepting skill banks \citep{cass}. Beta Shapley weights marginal contributions across coalition sizes, while Data Banzhaf studies robustness to noisy utility measurements \citep{betashapley,databanzhaf}. \method{} compares equal-capacity replacements to identify a first-order value projection aligned with the deployment constraint. Higher-order interactions and finite-sample errors can still impair selection.

\paragraph{Causal skill curation.}
ASSAY uses randomized masking to estimate skill--task effects for library restructuring and task-conditioned masking \citep{assay}. CaSKG calibrates skill relations through textual counterfactual probes for task-conditioned retrieval \citep{caskg}. \method{} studies skill evolution in a persistent textual skill. Its exchanges hold background rules fixed and estimate effects of skill replacement under a specified case and background distribution. Least-squares recovery from paired observations follows classical graph ranking \citep{hodgerank}; we connect this estimator to constrained skill selection and its decision errors.

\paragraph{Health agent benchmarks.}
MedCalc-Bench and its Verified repository provide numerical medical-calculation records and automatically scored feedback \citep{medcalc,medcalcverified}. This permits evaluation of skill evolution through accuracy, regressions, and task-specific behavior. MedAgentBench evaluates interaction with a virtual electronic health record and clinical APIs \citep{medagentbench}. Our static calculation study addresses one component of health-agent adaptation; interactive workflows and longitudinal validation remain open.

\section{Problem formulation}
\label{sec:formulation}

\paragraph{Skill definition.}
A rule is a short, reusable instruction clause. The agent's skill is an ordered rendering of its selected rules. A skill evolution step selects the next skill using externally supplied labeled feedback.

Let $f_\theta$ be a frozen language model, $M_t$ its current skill, and $x$ a task input with reference answer $y$. For a skill $S$, let
\begin{equation}
Y(x,S,\xi)=r\bigl(f_\theta(x,\operatorname{render}(S);\xi),y\bigr)\in[0,1],
\qquad V_t(S)=\E_{x\sim P_t,\xi}[Y(x,S,\xi)].
\label{eq:value}
\end{equation}
Here $\xi$ includes model and execution randomness; rendering fixes the order and formatting of rules. The value function is conditional on a fixed execution configuration, including reasoning mode and decoding settings, whose notation we suppress. The distribution $P_t$ specifies which health tasks matter at round $t$. We analyze each evolution step under its current task distribution and execution configuration.

Given a new labeled experience $e_t$, a proposal operator generates text $G_\theta(M_t,e_t)$. The controller constructs a candidate collection $D_t$ and a feasible family $\Omega_t$, then chooses $M_{t+1}\in\Omega_t$. The separation is substantive: the model determines what can be proposed, while the experimental controller determines what the available observations support selecting.

\subsection{Incumbent feasibility}
\label{sec:support}
Retention requires candidate support: keeping a previous rule requires that rule or an equivalent replacement to be available in $D_t$. To preserve the entire incumbent skill as a feasible choice, we impose
\begin{equation}
M_t\in\Omega_t\quad\text{for every non-initial round}.
\label{eq:incumbent}
\end{equation}
For the prototype, $|M_t|=k=4$, $G_\theta$ returns two new rules, $D_t=M_t\cup G_\theta(M_t,e_t)$ has $n=6$ distinct clauses, and $\Omega_t=\{S\subseteq D_t:|S|=k\}$. The first round generates six rules from an empty skill. Invalid or textually duplicate proposals retain the current skill.

\subsection{Value projection}
\label{sec:projection}
Condition on a fixed round, candidate collection, and value function, and suppress $t$. Let $Z\in\{0,1\}^n$ be the indicator of a uniformly chosen member of $\Omega_k$. Define the best first-order approximation
\begin{equation}
(a^*,b^*)=\arg\min_{a,b:\one^\top b=0}
\E_{Z\sim\operatorname{Unif}(\Omega_k)}\left[V(Z)-a-b^\top Z\right]^2.
\label{eq:projection}
\end{equation}
The zero-sum condition resolves the intercept ambiguity on a fixed-size slice. The projection applies to arbitrary skill-value functions, including those with interactions between rules. Selecting the largest $k$ coordinates optimizes the fitted first-order surrogate; nonlinear complementarity can make that selection suboptimal for $V$.

For $i\ne j$, sample a background $B$ uniformly from the $(k-1)$-subsets excluding $i,j$. The two treatments are $B\cup\{i\}$ and $B\cup\{j\}$. They have equal rule counts and share all other rules. Define
\begin{equation}
\delta_{ij}=\E_B\left[V(B\cup\{i\})-V(B\cup\{j\})\right].
\label{eq:exchange}
\end{equation}
\begin{proposition}[Fixed-capacity exchange identity]
\label{prop:identity}
For any real-valued $V$ on $\Omega_k$, with $1\le k<n$, the projection in \eqref{eq:projection} satisfies
\begin{equation}
b_i^*=\frac{n(n-1)}{k(n-k)}\operatorname{Cov}(Z_i,V(Z)),
\qquad \delta_{ij}=b_i^*-b_j^*.
\label{eq:identity}
\end{equation}
\end{proposition}
The result follows from the covariance matrix of fixed-size sampling and is proved in Appendix~\ref{app:proofs}. Equation~\eqref{eq:identity} averages over eligible backgrounds; individual-background contrasts can differ from this average. Any component $h(|S|)$ depending only on rule count cancels exactly in these exchanges. Token count, rule order, and meaning can still affect the measured contrast.

\section{Method}
\label{sec:method}
\subsection{Evidence collection on a comparison graph}
\label{sec:graph}
Direct evaluation estimates $V(S)$ separately for each feasible skill: the prototype has $\binom{6}{4}=15$ such sets. \method{} collects paired exchanges from \eqref{eq:exchange} to fit $n$ relative rule values, then selects $k$ rules by their summed value. The connected comparison graph places all rules on a common scale; the resulting first-order representation supports selection across the feasible family. For each directed edge $e=(i,j)$, we run the agent twice on the same feedback case with the two skills from \eqref{eq:exchange}. The measured difference is
\begin{equation}
d_e=Y(x_e,B_e+i,\xi_e^+)-Y(x_e,B_e+j,\xi_e^-).
\label{eq:observed}
\end{equation}
Let $A$ be the edge--vertex incidence matrix, with a $+1$ at $i$ and $-1$ at $j$. We estimate
\begin{equation}
\widehat b=\arg\min_{b:\one^\top b=0}\|d-Ab\|_2^2
=(A^\top A)^+A^\top d.
\label{eq:fit}
\end{equation}
Under the randomized backgrounds and a common case distribution, $\E[d\mid A]=Ab^*$ and hence $\E[\widehat b\mid A]=b^*$. This mean identity also holds for dependent edges; variance analysis additionally requires their covariance structure.

The prototype uses a randomly relabeled six-vertex cycle: six paired comparisons require 12 actor calls per evolution step, or 288 across 24 evolution steps, with each of three feedback cases serving two randomly assigned edges. Appendix~\ref{app:proofs} details exact cycle computation and its design rationale under independent, equal-variance noise; shared cases and unequal noise reduce that idealization.

\subsection{Skill evolution}
The next skill maximizes the fitted value, breaking exact ties by retaining more incumbent rules:
\begin{equation}
\widehat S\in\operatorname*{lexmax}_{S\in\Omega_k}
\left(\sum_{i\in S}\widehat b_i,\ |S\cap M_t|\right).
\label{eq:selection}
\end{equation}
A fixed random order resolves remaining ties; if every observed difference is zero, the incumbent is retained. Misleading score differences can still cause a harmful evolution step.

Algorithm~\ref{alg:ise} in Appendix~\ref{app:algorithm} gives the pseudocode for one evolution step; the actor and proposal templates appear in Figures~\ref{fig:actor-prompt} and~\ref{fig:proposal-prompt}.

\subsection{Decision error}
\label{sec:bound}
Let $\Omega\subseteq\Omega^*$, where $\Omega^*$ is a specified reference family and $\Omega$ is the available fixed-capacity family. Define the support gap $g=\max_{S\in\Omega^*}V(S)-\max_{S\in\Omega}V(S)$, approximation error $\epsilon=\max_{S\in\Omega}|V(S)-a^*-\sum_{i\in S}b_i^*|$, and coefficient error $\eta=\|\widehat b-b^*\|_\infty$.
\begin{proposition}[Support, approximation, and estimation]
\label{prop:bound}
The selection in \eqref{eq:selection} obeys
\begin{equation}
\max_{S\in\Omega^*}V(S)-V(\widehat S)
\le g+2\epsilon+2\min(k,n-k)\eta.
\label{eq:bound}
\end{equation}
If $M_t\in\Omega$, then $V(M_t)-V(\widehat S)\le2\epsilon+2\min(k,n-k)\eta$.
\end{proposition}
Support and approximation errors persist even with exact coefficient estimates, and numerical certification would require reliable bounds on each term. Appendix~\ref{app:proofs} gives the proof and a complementarity counterexample.

\section{Health calculation experiments}
\label{sec:experiments}
\subsection{Experimental setup}
The main experiment uses six MedCalc-Bench-Verified tasks in the order SIRS, HEART, PERC, CURB-65, Child-Pugh, and Cockcroft--Gault. A shared 96-case learning pool supplies 24 teaching and 72 feedback cases to \method{} and 72 training and 24 internal-validation cases to SkillOpt.

Comparison with no skill measures the change in performance from supplying learned rules. Comparison with SkillOpt assesses the final skills produced by two learning procedures. The separate shuffled-credit ablation tests the role of assigning estimated credits to their associated rules within the exchange controller.

We compare \method{} with SkillOpt \citep{skillopt} and the no-skill baseline. \method{} uses the six-cycle exchange design of Section~\ref{sec:graph}, with four sequential steps per task. SkillOpt optimizes persistent textual skills for a frozen actor. We adapt its official ReflACTTrainer to the common task and scoring interface, using four epochs with six 12-case batches per epoch and its internal-validation-selected state (step 5). No skill uses the common task instruction with an empty skill. Appendix~\ref{app:baselines} provides baseline implementation details.

The test panel contains 120 cases, 20 per task, sampled independently of model correctness. All cases originate from the source training split, with learning and evaluation roles assigned for this study. Record-identifier and text-overlap checks separate the test panel from the 96 learning cases, the 24-case development panel, and a separate 115-case evaluation panel. Appendix~\ref{app:main-protocol} details the filtering and sampling. These checks operate at the record and text levels; patient-level independence remains unverified.

Both methods learn with low reasoning and frozen model weights, with output caps of 4,096 tokens for the actor and 8,192 tokens for proposals and optimization. We retain low reasoning during learning to support problem solving and the construction of reusable rules. Optimization budgets and skill lengths differ between methods.

\subsection{Evaluation protocol}
All skills and evaluated checkpoints are fixed before test evaluation. Reported results use thinking-off inference with the same DeepSeek V4 Flash 0731 actor, temperature 0, and a 16,384-token output cap. Each case receives eight independent, interleaved state evaluations: no skill, \method{} at steps 4, 8, 12, 16, 20, and 24, and final SkillOpt. Case and within-case state order are randomized.

We disable thinking at evaluation to prioritize stable completion and inference efficiency. This choice was motivated by reasoning-enabled runs that could exhaust the available token budget. Each learned skill is frozen and supplied as instruction text to the same thinking-off actor, so the final comparison measures its utility under a common inference configuration. Calculation criteria, units, and procedural steps remain available through the prompt in either mode, although their effect can depend on the reasoning setting. Selection and final evaluation therefore measure skill value under their respective execution configurations, as formalized in Section~\ref{sec:formulation}.

The primary endpoint is equal-weight task-macro accuracy, equal to case accuracy on this balanced panel. Unparsed answers, communication failures, and non-normal completions count as incorrect. Integer tasks use Python rounding against the reference answer; Cockcroft--Gault uses the inclusive reference interval. Secondary diagnostics measure parser-based output validity and pooled accuracy among valid outputs (Appendix~\ref{app:output-quality}). The 95\% intervals use 20,000 task-stratified paired bootstrap replicates. Final-state comparisons use exact two-sided McNemar tests with Holm correction across three comparisons. Intervals are unadjusted and conditional on the frozen skills and one training trajectory per method; they do not account for training variability or method-development choices.

\subsection{Main results}
Table~\ref{tab:results} summarizes overall accuracy and output quality on the balanced 120-case panel. \method{} answers 72 cases correctly (60.00\%), compared with 67 for SkillOpt (55.83\%) and 62 without skills (51.67\%).

\begin{table}[!htbp]
\centering
\caption{Main results on 120 held-out cases with thinking-off inference. Accuracy gives equal weight to the six tasks; invalid outputs count as incorrect. Valid-only accuracy is a secondary diagnostic with method-specific denominators. Bold marks the highest observed overall accuracy.}
\label{tab:results}
\small
\begin{tabular}{lrrrr}
\toprule
Method & \shortstack[r]{Accuracy\\(\%)} & \shortstack[r]{Accuracy\\95\% CI} & \shortstack[r]{Valid outputs\\(\%)} & \shortstack[r]{Accuracy on\\valid outputs (\%)}\\
\midrule
No skill & 51.67 & [43.33, 60.00] & 100.00 & 51.67\\
SkillOpt & 55.83 & [47.50, 64.17] & 100.00 & 55.83\\
\midrule
\method{} & \textbf{60.00} & [52.50, 67.50] & 99.17 & 60.50\\
\bottomrule
\end{tabular}
\end{table}

\paragraph{Paired outcomes.}
The higher aggregate score combines corrections with losses on other cases. Relative to no skill, \method{} repairs 22 errors but loses 12 previously correct answers; relative to SkillOpt, it repairs 21 and loses 16. The paired 95\% intervals are $[-0.83,17.50]$ points against no skill and $[-5.00,13.33]$ against SkillOpt. Neither comparison is significant after Holm correction (Appendix~\ref{app:paired-results}).

\paragraph{Output quality.}
Both baselines return 120 valid outputs and \method{} returns 119, with one invalid response on Child-Pugh; across all eight states, 959 of 960 outputs are valid and none is truncated. The remaining errors are valid numerical answers that disagree with the reference: 47 for \method{}, 53 for SkillOpt, and 58 for no skill. Appendix~\ref{app:output-quality} specifies the accepted formats.

\FloatBarrier
\subsection{Task heterogeneity}
\begin{figure}[!t]
\centering
\includegraphics[width=\linewidth]{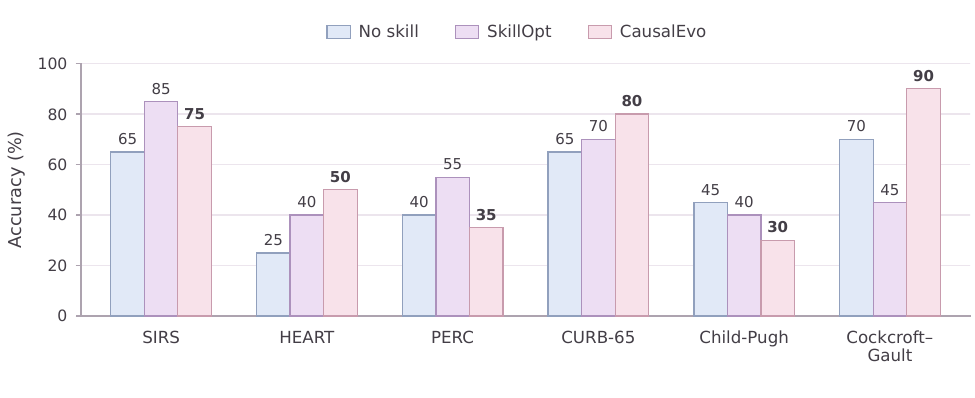}
\caption{Final task accuracy for the three frozen skill states. Each task has 20 held-out cases, so five percentage points equal one case. Invalid outputs count as incorrect. Exact values appear in Appendix~\ref{app:task-results}.}
\label{fig:task-results}
\end{figure}
Figure~\ref{fig:task-results} shows where the aggregate gains arise. \method{} leads SkillOpt on three of six tasks, but Cockcroft--Gault alone contributes nine additional correct answers, exceeding the overall five-case advantage. The other five tasks therefore contribute a net loss of four. SkillOpt leads on SIRS and PERC, while no skill performs best on Child-Pugh. The higher macro score thus combines task-specific gains and regressions.

\paragraph{Skill content and task behavior.}
Two of the four retained rules explicitly target Cockcroft--Gault, covering weight calculation and conversion precision (Figure~\ref{fig:final-skill}). This concentration of task-specific instructions accompanies the largest observed task-level gain. The other two rules address threshold logic and measurement timing; one applies its instruction to any scoring system with numeric thresholds. All four rules are supplied to every task, so the final skill combines uneven calculator-specific coverage with broadly scoped instructions.

\Needspace{0.4\textheight}
\subsection{Learning dynamics}
\begin{wrapfigure}{r}{0.48\textwidth}
\centering
\includegraphics[width=\linewidth]{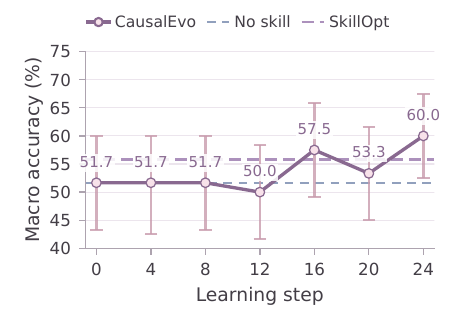}
\caption{Macro accuracy over learning steps on the fixed 120-case test panel. Step 0 uses an empty skill. Error bars are conditional 95\% task-stratified bootstrap intervals. Dashed lines show no skill and the selected SkillOpt endpoint as fixed references.}
\label{fig:macro-learning}
\end{wrapfigure}
\paragraph{Skill changes.}
All 24 proposal rounds are valid. After initialization, 12 of 23 rounds change the skill and 11 retain all four incumbent rules. The final skill contains 180 whitespace words, compared with 999 for SkillOpt, approximately 82\% fewer. Figure~\ref{fig:final-skill} reproduces it in full. Section~\ref{sec:training-tokens} reports training token use under these different skill representations and optimization budgets.

\paragraph{Accuracy over learning.}
Figure~\ref{fig:macro-learning} follows \method{} on the same 120 cases throughout learning. Accuracy stays at baseline through step 8, dips at step 12, and then rises, with an intervening decline at step 20, to 60.00\% at the prescribed endpoint of step 24.

\paragraph{Task transfer.}
The final four steps add eight correct answers: three on the currently taught Cockcroft--Gault task and a net five across earlier tasks, as HEART rises from 35\% to 50\% and PERC from 25\% to 35\%. Losses also occur: Child-Pugh falls from 50\% at step 16 to 30\% at the endpoint. Average immediate learning acquisition is $-3.33$ points (95\% interval $[-12.50,5.83]$), while backward and forward transfer are both $+7.00$ points, with intervals $[-3.00,17.00]$ and $[-4.00,18.00]$ (definitions in Appendix~\ref{app:main-protocol}). All three intervals include zero: the endpoint gain coexists with weak immediate acquisition and an uncertain direction of transfer.

\FloatBarrier
\subsection{Credit-assignment ablation}
\label{sec:ablation}
Table~\ref{tab:ablation-results} isolates credit--rule correspondence within the controller. This ablation uses a separate dataset with a relatively small test set of 24 cases, four per task. The control permutes the six fitted credits before selection while retaining the proposal interface, candidate construction, six paired exchanges, and tie rule; each trajectory subsequently proposes from its own skill. Training uses 24 teaching and 72 feedback cases.

\begin{table}[H]
\centering
\caption{Credit-assignment ablation on a separate dataset with a 24-case test set. Phase mean averages the six phase-end seen-task accuracies. Bold marks the highest observed value in each column.}
\label{tab:ablation-results}
\begin{tabular}{lrr}
\toprule
Setting & Final accuracy (\%) & Phase mean (\%)\\
\midrule
No skill & 62.50 & 61.94\\
Shuffled credit & 62.50 & 68.68\\
\midrule
\method{} & \textbf{70.83} & \textbf{73.33}\\
\bottomrule
\end{tabular}
\end{table}

Intact credits yield 17 correct answers, compared with 15 for each control, and a 4.65-point phase-mean gain over shuffled credit. The final paired gain is $+8.33$ points (95\% interval $[-4.17,25.00]$). These estimates condition on single trajectories and a panel reused for method development. Appendix~\ref{app:ablation} reports the protocol, phase curves, and failure diagnostics.

\subsection{Training token use}
\label{sec:training-tokens}
\begin{figure}[h]
\centering
\includegraphics[width=0.55\linewidth]{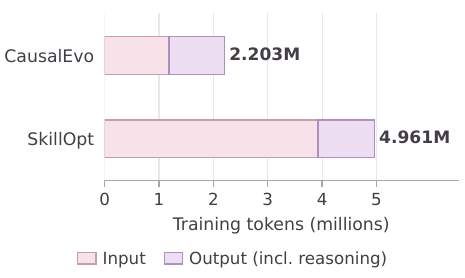}
\caption{Training token use in the recorded configurations, whose optimization budgets differ; output includes reasoning. Exact counts are in Appendix~\ref{app:training-tokens}.}
\label{fig:training-tokens}
\end{figure}

Across the recorded training configurations with different optimization budgets, \method{} uses 2.203 million training tokens, compared with 4.961 million for SkillOpt: a \textbf{55.6\% reduction} (Figure~\ref{fig:training-tokens}). Input tokens fall by 69.8\%, while output tokens, including reasoning, fall by 1.9\%. The saving therefore comes predominantly from lower aggregate input volume.

The exchange design specifies 12 scoring calls per evolution step. Including one teaching call and one proposal call gives 14 requests per step and 336 over 24 steps, compared with SkillOpt's 839 training requests. Token volume also depends on prompt content, skill length, and generated responses. The reported totals combine these factors with each method's optimization schedule; the learned skills contain 180 words for \method{} and 999 for SkillOpt. Call counts describe the work performed, while the input/output breakdown identifies where the token difference occurs.

\section{Discussion}
\label{sec:discussion}
\method{} makes the evidence for reusing a health-agent skill explicit: which candidate skills were tested, under what capacity constraint, and with which measured outcomes. Medical calculations provide an inspectable setting for this mechanism, while the four-rule prototype keeps individual evolution steps small enough to examine. Proposition~\ref{prop:bound} separates three improvement targets: candidate coverage, interactions between rules, and estimation noise.

The empirical findings also inform reliability evaluation. Aggregate gains coexist with case-level regressions. Nearly universal output validity leaves substantial numerical error. Evaluating an adaptive health agent therefore requires task-level accuracy and measures of which correct answers become errors after a skill evolution step.

A concrete extension is electronic-record querying with clinical tools \citep{medagentbench}. Rules could specify a retrieval time window, a unit check, or a tool-call sequence. Each paired evaluation would start from the same user request and record snapshot, share $k-1$ rules, and exchange the remaining rule. A reference answer or predefined task-completion criterion would score the resulting trajectory. Resetting the environment between runs would prevent one run's tool actions from changing the next run's starting conditions. In this mapping, the feedback case includes the query and initial environment, and the reward measures the outcome of a multi-step interaction.

Persistent rules also require domain review before deployment: a retained rule is a standing instruction to the model, and the controller judges it only by measured accuracy on the available cases. The four-rule capacity is a design choice; larger skills need more exchanges for a connected comparison graph, and the trade-off between capacity, call budget, and estimation noise remains to be characterized.

\section{Limitations}
\label{sec:limitations}
The evidence covers 120 static medical-calculation cases from six previously represented calculators and one training trajectory per method; training backends, skill lengths, and optimization budgets differ. Broader comparator coverage with matched budgets, independent learning runs, an isolated test of incumbent preservation, a comparison against exhaustive evaluation of all feasible skills, and evaluation on new calculators would strengthen the evidence. The credit-assignment ablation reused a development panel. Selection can overfit its three-case feedback panel and incurs projection error when rules interact.

Some development labels depend on disputed choices of measurement time point or units; scores measure agreement with the frozen references. Erroneous labels can reinforce an incorrect rule and create unwarranted confidence. Clinical application would require independent validation and oversight appropriate to its intended use.

\section{Conclusion}
\method{} performs causal skill self-evolution for health LLM agents: a frozen model proposes rules, paired fixed-capacity exchanges estimate their relative values, and selection keeps the incumbent skill available at every step. On 120 held-out medical-calculation cases it reaches 60.00\% accuracy; paired differences remain uncertain and task-level gains vary. Across the recorded training configurations, which use different optimization budgets, it consumes 55.6\% fewer training tokens than SkillOpt, primarily through lower input volume. The method turns feedback into explicit, inspectable decisions about retaining or replacing reusable skills.

\FloatBarrier
\bibliographystyle{plainnat}
\bibliography{references}

\newpage
\appendix
\raggedbottom
\section{Proofs}
\label{app:proofs}
All statements condition on a fixed candidate collection and a fixed value function on its size-$k$ subsets. An index denotes a particular rule string with a deterministic rendering convention. The intervention includes the effect of inserting that text into the resulting prompt. Semantically similar strings may have different effects.

\subsection{Exchange identity}
For a uniform size-$k$ indicator vector, $\E[Z_i]=k/n$,
\begin{equation}
\operatorname{Var}(Z_i)=\frac{k(n-k)}{n^2},\qquad
\operatorname{Cov}(Z_i,Z_j)=-\frac{k(n-k)}{n^2(n-1)}\quad(i\ne j).
\label{eq:covariance}
\end{equation}
The covariance matrix therefore acts as multiplication by $q=k(n-k)/[n(n-1)]$ on the subspace $\one^\perp$. On that subspace, the normal equations of the least-squares projection give $q b_i^*=\operatorname{Cov}(Z_i,V(Z))$. The optimal intercept is $a^*=\E[V(Z)]$, because $\E[(b^*)^\top Z]=0$. This proves the first identity and the uniqueness of the zero-sum coefficient vector.

For the second identity, write
\begin{align}
\operatorname{Cov}(Z_i,V)-\operatorname{Cov}(Z_j,V)
 &=\E[(Z_i-Z_j)V] \nonumber\\
 &=q\left(\E[V\mid Z_i=1,Z_j=0]-\E[V\mid Z_i=0,Z_j=1]\right)
 =q\delta_{ij}.
\label{eq:exchangeproof}
\end{align}
The two conditional backgrounds are uniform $(k-1)$-subsets of the same $n-2$ remaining indices. Dividing by $q$ proves the result for arbitrary $V$. The derivation requires uniform averaging over these backgrounds.\hfill$\square$

\subsection{Graph estimation}
For a connected graph, $L=A^\top A$ has nullspace $\operatorname{span}(\one)$. If the conditional mean of each observed edge difference is the corresponding $b_i^*-b_j^*$, then
\begin{equation}
\E[\widehat b\mid A]=L^+A^\top A b^*=b^*,
\label{eq:unbiased}
\end{equation}
since $b^*\in\one^\perp$. With independent homoskedastic edge noise of variance $\sigma^2$, the covariance on this subspace is $\sigma^2 L^+$. Correlated or unequal-variance errors require a corresponding covariance adjustment.

For a six-cycle with vertices $v_0,\ldots,v_5$ and oriented differences $d_0,\ldots,d_5$, let $\bar d=\sum_e d_e/6$, set $u_{v_0}=0$, and recursively set $u_{v_{e+1}}=u_{v_e}-(d_e-\bar d)$. Centering the six $u$ values gives \eqref{eq:fit}. The fitted differences are $d_e-\bar d$, and the residual sum of squares is $6\bar d^2$. A zero observed residual is compatible with nonlinear rule interactions, because the mean identity itself holds for arbitrary $V$.

By the matrix-tree theorem, the product of the nonzero Laplacian eigenvalues is $n$ times the number of spanning trees. A connected simple graph on six vertices with six edges is unicyclic. If its unique cycle has length $\ell$, it has $\ell$ spanning trees: delete one of the cycle edges, while every attached tree edge must remain. Thus $\ell\le6$, with equality for the six-cycle. This proves D-optimality only among this restricted family under the equal-variance independent-noise model. In our experiment, two edges share each feedback case, and different model responses can have different variability.

\subsection{Decision bound}
Let $S^*$ maximize $V$ within $\Omega$. The uniform approximation bound gives
\begin{equation}
V(S^*)-V(\widehat S)
\le 2\epsilon+\sum_{i\in S^*}b_i^*-\sum_{i\in\widehat S}b_i^*.
\label{eq:boundproof1}
\end{equation}
Because $\widehat S$ maximizes the estimated coefficient sum, adding and subtracting $\widehat b$ bounds the final two sums by
\begin{equation}
\sum_{i\in S^*\setminus\widehat S}(b_i^*-\widehat b_i)
-\sum_{i\in\widehat S\setminus S^*}(b_i^*-\widehat b_i)
\le |S^*\mathbin{\triangle}\widehat S|\eta
\le 2\min(k,n-k)\eta.
\label{eq:boundproof2}
\end{equation}
The last inequality follows because both sets have size $k$ in a universe of size $n$. Adding the support gap proves \eqref{eq:bound}. If $M_t\in\Omega$, then $V(M_t)\le V(S^*)$, giving the incumbent comparison. Lexicographic tie-breaking preserves the first objective, so the proof applies.\hfill$\square$

\subsection{Complementarity counterexample}
Take $n=4,k=2$ and assign values
\begin{equation}
V(12)=1,\quad V(13)=V(14)=0.6,\quad V(23)=V(24)=0,\quad V(34)=0.9.
\label{eq:counterexample}
\end{equation}
The exact zero-sum projection coefficients are $(13,-11,-1,-1)/40$. Selecting the two largest coefficients returns $\{1,3\}$ or $\{1,4\}$ with value $0.6$, whereas the optimal feasible skill $\{1,2\}$ has value $1$. Approximation error therefore remains even with exact coefficient estimates.

The accompanying mathematical checker verifies the exchange identity on arbitrary finite value tables, the decision-loss bound, this counterexample, and the restricted six-cycle design calculation.

\section{Implementation details}
\label{app:implementation}
This section documents the actor and proposal interfaces, the training configuration, and the pseudocode shared by the main experiment and the credit-assignment ablation.

\subsection{Model interfaces}
Figures~\ref{fig:actor-prompt} and~\ref{fig:proposal-prompt} show the actor and proposal templates. Braced fields describe the variable inputs schematically; the fixed instructions retain their documented wording.

\begin{figure}[H]
\centering
\begin{promptbox}{Prompt Template for the Calculation Actor}
\promptfield{System}
Solve the requested medical calculation as a research exercise using the supplied patient note and calculator name. Show a brief calculation (maximum160 words), then end with a separate line exactly FINAL: followed by one numeric value only, with no units or commas. Do not provide treatment advice. Do not output JSON.

\promptfield{Selected Rules -- appended to System}
\promptslot{selected\_rules\_in\_rendering\_order}
\promptdivider
\promptfield{User Input}
\promptslot{patient\_note}\par
\promptslot{calculation\_question}
\end{promptbox}
\caption{Actor prompt. Selected rules follow the fixed instruction; the no-skill state leaves that field empty. Field headings indicate message structure.}
\label{fig:actor-prompt}
\end{figure}

The user message contains the original patient note and question. An output is valid only if it ends normally and the fixed numerical parser extracts an explicit answer. Integer tasks require \texttt{round(value)} to equal the numerical gold answer, using Python rounding. Decimal tasks require the value to lie within the source's inclusive lower and upper bounds.

\begin{figure}[H]
\centering
\begin{promptbox}{Prompt Template for Skill Proposals}
\promptfield{System}
Propose exactly \promptslot{m} reusable medical-calculation rules from the current rules and labeled attempt.\par
Each rule must state its applicability and procedure in 5--80 English words; propose additions, corrections, or compact replacements.\par
Treat the example as evidence, not a universal condition; exclude patient values, narratives, and treatment advice.\par
Return only RULE 1: through RULE \promptslot{m}: on separate lines; the program preserves old rules and decides what to retain.
\promptdivider
\promptfield{Current Rules}
\promptslot{current\_rules}
\promptfield{Labeled Teaching Case}
\promptslot{calculator\_name, patient\_note, question}\par
\promptslot{reference\_answer, reference\_explanation}
\promptfield{Actor Attempt}
\promptslot{attempt\_text, correctness, validity, finish\_reason}
\end{promptbox}
\caption{Skill-proposal template, with $m=6$ at initialization and $m=2$ for a nonempty incumbent. The lower fields summarize the structured user payload. The program selects rules from the scored comparisons.}
\label{fig:proposal-prompt}
\end{figure}

The structured user payload contains current rules; the teaching case's calculator name, note, question, reference answer and explanation; and the actual attempt's text, correctness, validity, and finish reason. Evaluation cases remain separate from this payload. The parser requires ordered rule labels, distinct rule text, normal completion, and 5--80 English words per rule. Exact duplication of an incumbent clause invalidates the proposal; semantic near-duplication can pass this check. An invalid proposal leaves the incumbent skill unchanged and is recorded.

\subsection{Training configuration}
Table~\ref{tab:execution} lists the skill evolution settings. Each feedback case has equal probability of assignment to each edge. The estimator's mean concerns the realized three-case panel and its task composition; generalization to a broader health-task distribution requires additional evidence.

\begin{table}[ht]
\centering
\caption{Settings for skill evolution.}
\label{tab:execution}
\begin{tabularx}{\linewidth}{lX}
\toprule
Setting & Value\\
\midrule
Model identifier & \texttt{deepseek/deepseek-v4-flash-0731}\\
Temperature & Actor 0, proposer 0.3\\
Output caps & Actor 4,096; proposer 8,192 tokens\\
State constraint & Four rules, 5--80 words each\\
Comparison design & Random six-cycle; six pairs, 12 actor calls per valid evolution step\\
\bottomrule
\end{tabularx}
\end{table}

\subsection{Pseudocode}
\label{app:algorithm}
Algorithm~\ref{alg:ise} specifies one skill evolution step. Here $D[I]$ preserves increasing candidate-index order, $[0]=\varnothing$, and $Y$ is the scored actor in \eqref{eq:value}. The proposal-validity check follows the parser above; the ablation's random seed schedule is given in Appendix~\ref{app:protocol}. Invalid actor outputs have score zero.

\begin{algorithm}[H]
\small
\caption{\method{}: one skill evolution step}
\label{alg:ise}
\DontPrintSemicolon
\KwIn{Frozen actor $f_\theta$ and proposer $G_\theta$; skill $M_t$ with $|M_t|\in\{0,4\}$; labeled lesson $\ell_t$ with input $x_t$; labeled feedback cases $\mathcal F_t=(x_{t,1},x_{t,2},x_{t,3})$; random seed schedule $s_t$.}
\KwOut{Next skill $M_{t+1}$.}
$k\gets4$; $n\gets6$; $m\gets n-|M_t|$\;
$o_t\gets f_\theta(x_t,\operatorname{render}(M_t))$\;
$U_t\gets G_\theta(M_t,\ell_t,o_t;m)$ \tcp*{Request $m$ candidate rules}
\If{$U_t$ is invalid \textbf{or} $|\operatorname{unique}(M_t\mathbin\Vert U_t)|\ne n$}{
    \Return{$M_t$}\;
}
$D\gets M_t\mathbin\Vert U_t$; $I_{\mathrm{old}}\gets[|M_t|]$; $\mathcal Q\gets\varnothing$\;
$\pi\sim\operatorname{Unif}(\mathfrak S_n)$; $\pi_{n+1}\gets\pi_1$\;
$E\gets\bigl((\pi_e,\pi_{e+1})\bigr)_{e=1}^{n}$ \tcp*{Directed cycle}
$c\gets\operatorname{Permute}(1,1,2,2,3,3)$ \tcp*{Two edges per case}
\For{$e=1,\ldots,n$}{
    $(i,j)\gets E_e$\;
    $B_e\sim\operatorname{Unif}\bigl(\{B\subseteq[n]\setminus\{i,j\}:|B|=k-1\}\bigr)$\;
    $S_e^+\gets D[B_e\cup\{i\}]$; $S_e^-\gets D[B_e\cup\{j\}]$\;
    $\mathcal Q\gets\mathcal Q\cup\{(e,+),(e,-)\}$\;
}
\ForEach{$(e,\sigma)\in\operatorname{Permute}(\mathcal Q)$}{
    $z_e^\sigma\gets Y(x_{t,c_e},S_e^\sigma,\xi_e^\sigma)$\;
}
$d\gets(z_e^+-z_e^-)_{e=1}^{n}$; $A\gets\operatorname{Incidence}(E)$\;
$\widehat b\gets(A^\top A)^+A^\top d$ \tcp*{Exact cycle fit; \eqref{eq:fit}}
$\mathcal I^\star\gets\underset{I\subseteq[n],\,|I|=k}{\operatorname{arg\,lexmax}}\left(\sum_{i\in I}\widehat b_i,\ |I\cap I_{\mathrm{old}}|\right)$\;
$I^\star\gets\operatorname{TieBreak}(\mathcal I^\star;s_t)$ \tcp*{Fixed candidate order}
$M_{t+1}\gets D[I^\star]$\;
\Return{$M_{t+1}$}\;
\end{algorithm}

The manuscript source includes figure-generation code, a mathematical checker, aggregate results, and record-role identifiers. Complete training code and raw responses are not currently included in the package.

\section{Main experiment: supplementary protocol}
\label{app:main-protocol}
The test panel is drawn from the frozen MedCalc-Bench-Verified source training split. Child-Pugh uses Synthetic cases and the other tasks use Extracted cases, following the source labels. Overlap filtering checks record identifiers, normalized text, complete-case matches against auxiliary evaluation records, and five-word shingle Jaccard similarity with threshold 0.85. Of 824 candidates, 592 pass these checks before sampling 20 cases per task.

Test selection uses seed 2026091001, schedule randomization uses 2026091011, and bootstrap resampling uses 2026091012. A bootstrap draw samples 20 cases with replacement within each task, carrying all state responses together. Exact McNemar tests use discordant pairs, and the three final-state tests form one Holm family. This correction applies only to the three stated comparisons.

Let $A_{t,j}$ be accuracy on task $j$ after stage $t$, where $t=0$ denotes no skill and stages 1--6 denote steps 4--24. All entries use the same 20 test cases per task. We report
\begin{align*}
\mathrm{ACC}&=\tfrac16\sum_{j=1}^6 A_{6,j}, &
\mathrm{RG}&=\tfrac16\sum_{j=1}^6(A_{6,j}-A_{0,j}),\\
\mathrm{LA}&=\tfrac16\sum_{j=1}^6(A_{j,j}-A_{j-1,j}), &
\mathrm{BWT}&=\tfrac15\sum_{j=1}^5(A_{6,j}-A_{j,j}),\\
\mathrm{FWT}&=\tfrac15\sum_{j=2}^6(A_{j-1,j}-A_{0,j}).
\end{align*}
SkillOpt uses mixed-task optimization and contributes its selected endpoint to the final comparisons. The sequential learning matrix tracks \method{}. Training uses low reasoning; endpoint and checkpoint evaluation uses reasoning disabled. Each round runs six paired exchanges on three feedback cases, fits relative rule values, and selects the top four rules.

\subsection{Baselines}
\label{app:baselines}
The SkillOpt comparison evaluates final skills learned from a shared case pool through a common frozen-actor interface. We adapt the released ReflACTTrainer using its environment interface for new tasks. Training budgets and optimization schedules differ between methods. The four-rule capacity applies to \method{}'s candidate interventions; SkillOpt retains its own text representation.

Other skill-evolution systems were not run for two reasons. At the time of writing, SkillCAT and CASS have no public code, and the CaSKG release omits its skill corpus and experiment workspace \citep{skillcat,cass,caskg}. Independently of code, each system requires protocol choices that our setup does not specify: SkillCAT and SkillBoost need multi-trajectory collection, replay budgets, and inference-time routing \citep{skillcat,skillboost}; SkillShapley values steps within an existing skill and needs a proposal and evolution policy to become a learning baseline \citep{skillshapley}; ASSAY and CaSKG need task-conditioned masking or retrieval \citep{assay,caskg}. Established optimizers such as ACE, GEPA, and TextGrad remain relevant controls \citep{ace,gepa,textgrad}. Broader comparator coverage and training-budget control remain limitations of the present evaluation.

\subsection{Training token totals}
\label{app:training-tokens}
Table~\ref{tab:training-tokens} reports aggregate usage for skill learning and optimization, separately from fixed-skill test evaluation. Output counts include reasoning tokens. Total tokens equal input plus output tokens; request counts are reported separately. The comparison uses each method's recorded training configuration and optimization budget.

\begin{table}[ht]
\centering
\caption{Training usage for the main experiment.}
\label{tab:training-tokens}
\small
\begin{tabular}{lrrrr}
\toprule
Method & Requests & Input tokens & \shortstack[r]{Output tokens\\including reasoning} & Total tokens\\
\midrule
SkillOpt & 839 & 3,924,153 & 1,036,499 & 4,960,652\\
\method{} & 336 & 1,186,029 & 1,016,551 & \textbf{2,202,580}\\
\bottomrule
\end{tabular}
\end{table}

\subsection{Paired statistical results}
\label{app:paired-results}
\label{app:score-verification}
All 960 saved predictions were rescored offline with the fixed parser and reference labels, and the reported statistics were recomputed from those scores. The final skills answer 72 cases correctly for \method{}, 67 for SkillOpt, and 62 for no skill. Relative to SkillOpt, \method{} corrects 21 previously incorrect answers and changes 16 correct answers to incorrect ones. Its accuracy difference is $+4.17$ percentage points, with a conditional 95\% bootstrap interval of $[-5.00,13.33]$ and Holm-adjusted $p=0.9737$. Relative to no skill, it corrects 22 answers and worsens 12, giving $+8.33$ points, interval $[-0.83,17.50]$, and adjusted $p=0.3643$. SkillOpt corrects 19 answers and worsens 14 relative to no skill, giving $+4.17$ points, interval $[-5.00,13.33]$, and adjusted $p=0.9737$. No comparison is significant after correction. The intervals condition on the learned skills and do not capture training variability.

\subsection{Final task accuracies}
\label{app:task-results}
Table~\ref{tab:task-results} provides the values plotted in Figure~\ref{fig:task-results}. Each task contains 20 held-out cases, so one additional correct answer changes task accuracy by five percentage points. The single invalid final \method{} output is retained in the Child-Pugh denominator.

\begin{table}[ht]
\centering
\caption{Final accuracy (\%) on each task, with 20 held-out cases per task. Bold marks the highest observed accuracy in each column.}
\label{tab:task-results}
\small
\begin{tabular}{lrrrrrr}
\toprule
Method & SIRS & HEART & PERC & CURB-65 & Child-Pugh & \shortstack[r]{Cockcroft--\\Gault}\\
\midrule
No skill & 65 & 25 & 40 & 65 & \textbf{45} & 70\\
SkillOpt & \textbf{85} & 40 & \textbf{55} & 70 & 40 & 45\\
\midrule
\method{} & 75 & \textbf{50} & 35 & \textbf{80} & 30 & \textbf{90}\\
\bottomrule
\end{tabular}
\end{table}

\subsection{Output quality}
\label{app:output-quality}
An output is valid when its request succeeds, completion ends normally, and the fixed parser accepts one numeric answer. The parser accepts numeric-only responses and supported final-line labels, so validity measures acceptance across these formats. The aggregate records lack counts for each accepted format and the specific cause of the lone invalid output, preventing separate estimates of strict \texttt{FINAL:} compliance and parse-failure rate.

For each final state, let $N=120$, $V$ be the number of valid outputs, and $C$ the number of correct answers. Output validity is $V/N$, and valid-only accuracy is $C/V$. The latter pools the valid cases and can have a different task mixture and denominator for each method. Overall accuracy remains $C/N$. The three disjoint outcome categories in Table~\ref{tab:output-quality} sum to $N$: correct answers, valid but incorrect answers ($V-C$), and invalid outputs ($N-V$).

\begin{table}[ht]
\centering
\caption{Final-state output decomposition on the 120-case panel. All values are counts. Invalid outputs remain errors in the primary accuracy metric; truncation is one possible invalid-output cause.}
\label{tab:output-quality}
\small
\begin{tabular}{lrrrr}
\toprule
Method & Correct & \shortstack[r]{Valid but\\incorrect} & Invalid & \shortstack[r]{Length\\truncations}\\
\midrule
No skill & 62 & 58 & 0 & 0\\
SkillOpt & 67 & 53 & 0 & 0\\
\midrule
\method{} & 72 & 47 & 1 & 0\\
\bottomrule
\end{tabular}
\end{table}

Across all eight evaluated states, validity is 959 of 960 outputs and the recorded truncation rate is zero. This pool contains repeated evaluations of the same 120 cases. Only failures that return no model output are eligible for a retry. Completed answers, including formatting failures, are scored once. The evaluation truncation count covers endpoint and checkpoint responses; the three capped SkillOpt optimizer responses belong to training.

\subsection{A complete learned skill}
\label{app:skill-example}
Figure~\ref{fig:final-skill} reproduces the final skill learned in the main experiment. Two rules address thresholds and source measurements, and two specify Cockcroft--Gault procedures. The same four-rule skill is deployed across all six evaluation tasks.

\begin{figure}[H]
\centering
\begin{promptbox}{Final \method{} skill (step 24)}
\input{figures/final_skill_text}
\end{promptbox}
\caption{Complete deployed skill from the main experiment at step 24. The actor receives these four rules after its fixed task instruction. The rule text is preserved from the saved skill artifact.}
\label{fig:final-skill}
\end{figure}

\section{Credit-assignment ablation}
\label{app:ablation}
This ablation tests credit--rule correspondence within the \method{} controller. It uses a separate dataset with a 24-case test set.

\subsection{Experimental setup}
The six tasks are introduced in the same order as in the main experiment, with four learning rounds each. We compare two learning trajectories, \method{} and shuffled credit. Both use the model interfaces and training configuration of Appendix~\ref{app:implementation}, the same incumbent-containing feasible sets, intervention design, and tie rule. The control permutes the six fitted credits before assigning them to rules. When the trajectories enter a round with identical skills, they share teaching and proposal responses; intervention calls remain separate. After their states diverge, each trajectory proposes from its own skill. Truncated proposals leave the skill unchanged, and invalid actor outputs score zero.

\subsection{Dataset allocation and seeds}
\label{app:protocol}
We use MedCalc-Bench-Verified, available under the CC-BY-SA-4.0 license \citep{medcalcverified}, with a fixed development allocation and unchanged reference labels: 24 teaching cases, 72 feedback cases, and 24 development evaluation cases (four per task). These learning and development records are drawn from the source training split, with separation from each other and from the main test panel established at the note-ID and exact-text levels. Table~\ref{tab:allocation} lists the task composition. The learning allocation contains 88 Extracted and 8 Synthetic records; evaluation contains 20 Extracted and 4 Synthetic records, using the dataset's own note-type labels. The manuscript source includes a machine-readable allocation manifest containing record identifiers and roles.

\begin{table}[ht]
\centering
\caption{Fixed allocation. Each task is introduced for four consecutive rounds. Feedback can come from tasks introduced previously; early tasks consequently appear more often.}
\label{tab:allocation}
\begin{tabular}{llrrr}
\toprule
ID & Task & Teaching & Feedback & Development\\
\midrule
51 & SIRS & 4 & 32 & 4\\
18 & HEART & 4 & 18 & 4\\
48 & PERC & 4 & 10 & 4\\
45 & CURB-65 & 4 & 6 & 4\\
15 & Child-Pugh & 4 & 4 & 4\\
2 & Cockcroft--Gault & 4 & 2 & 4\\
\midrule
 & Total & 24 & 72 & 24\\
\bottomrule
\end{tabular}
\end{table}

For round index $t=0,\ldots,23$, the implementation uses seeds $181000+t$ for the cycle and backgrounds, $182000+t$ for intervention order, $183000+10t+e$ for the two executions on edge $e$, and $184000+t$ for final tie-breaking. The shuffled-credit permutation uses the last seed plus 100000. Teaching and proposal seeds are $125000+t$ and $126000+t$. The two policies use matched sampling schedules; stochastic model outputs may still differ.

\subsection{Evaluation protocol}
All learning states are finalized before evaluation begins. At the end of each four-round task phase, we evaluate every introduced task on its four development cases, yielding 84 predictions per trajectory. The empty skill is evaluated on the 24 cases once. The primary endpoint is final task-macro accuracy; with four cases per task, this equals final case accuracy. A secondary endpoint averages the six phase-end accuracies over the tasks introduced by each phase; that curve reflects both performance changes and the expanding task mixture, and empty-skill outcomes are reused to compute the corresponding changing-mixture reference.

Bootstrap intervals use 10,000 draws, sampling four paired cases with replacement independently within each of the six task strata, with random-number generator seed 188000. This analysis reuses the evaluation panel for method development and includes one learning trajectory per method; the intervals quantify case-sampling uncertainty conditional on the two realized trajectories and do not account for method-development choices.

\subsection{Results}
\begin{figure}[t]
\centering
\includegraphics[width=0.8\linewidth]{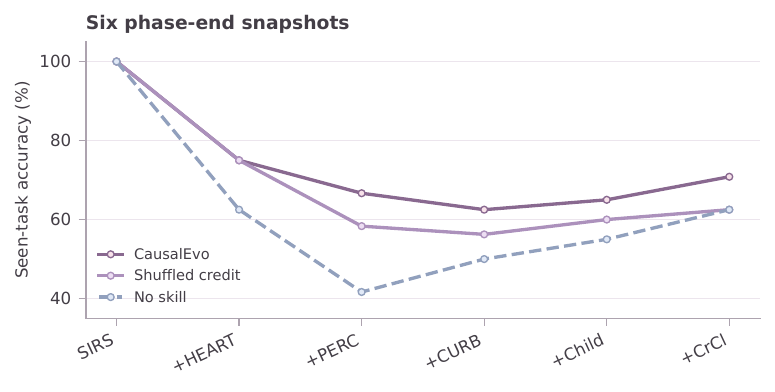}
\caption{Accuracy at each phase end on all tasks introduced so far. The evaluated task mixture expands with each phase.}
\label{fig:ablation-results}
\end{figure}

Table~\ref{tab:ablation-results} shows that \method{} answers 17 of 24 development cases correctly, compared with 15 for each control. Its phase-mean advantage over shuffled credit is 4.65 percentage points (Figure~\ref{fig:ablation-results}). The paired final difference from shuffled credit is $+8.33$ percentage points, with a conditional 95\% interval of $[-4.17,25.00]$. The difference from no skill is also $+8.33$ points, with interval $[0.00,16.67]$. All 24 final answers are valid for both learned trajectories; 23 are valid for no skill.

\subsection{Diagnostics}
\label{app:diagnostics}
\paragraph{Paired outcomes.}
Both gains over the empty skill occur on HEART, where both methods produced complete, valid answers. Relative to shuffled credit, the intact-credit trajectory gains one case each on HEART, Child-Pugh, and Cockcroft--Gault, and loses one PERC case. All four discordant pairs have valid outputs on both sides, so the differences reflect answer correctness; case identifiers are retained in the evidence files. In the PERC loss, the intact-credit trajectory returns a criterion count opposite to the reference, illustrating a task-level failure despite the aggregate gain.

\paragraph{Learning dynamics.}
\method{} makes 20 valid proposals in 24 rounds, changes its skill in 12 evolution steps, and encounters four truncated proposals. The shuffled control makes 23 valid proposals and changes its skill in 10 rounds. Across valid evolution steps, 18 of 120 exchange edges in \method{} and 13 of 138 in the control have nonzero binary differences; seven and thirteen steps, respectively, have all-zero edges. The measured distinctions between paired skill sets are therefore sparse.

\paragraph{Skill content.}
The final \method{} skill's four rules first appear in rounds 9, 22, 23, and 24. Two express near-duplicate body-weight procedures, one retains an editing reference to a rule number that can change after subset selection, and a unit-conversion rule includes a broadly worded time-point instruction. The control retains broader unit and source-value instructions. Both final documents provide incomplete explicit coverage of the six named calculation procedures, exposing limitations in semantic coverage and coherent composition. The last exchange graph has zero residual alongside the redundant final rules, illustrating the distinction between graph fit and skill-set quality.

\section{Use of large language models}
DeepSeek serves as the frozen actor and proposer. An LLM-based coding assistant supported implementation, mathematical exposition, literature organization, and drafting.

\end{document}